**Grammatical cues to subjecthood are redundant in a majority of simple clauses across languages**

Kyle Mahowald‡[1], Evgeniia Diachek‡[2], Edward Gibson[3], Evelina Fedorenko*[3,4] and Richard Futrell*[5]

‡ Co-first authors
* Co-senior authors

**Affiliations**

1. The University of Texas at Austin, Linguistics
2. Vanderbilt University, Psychology and Human Development
3. Massachusetts Institute of Technology, Brain and Cognitive Sciences
4. Massachusetts Institute of Technology, McGovern Institute for Brain Research
5. University of California, Irvine, Language Science

**Contributions:**
Conceptualization: KM, EG, EF, RF
Design and materials creation: KM, ED, EF, RF
Human data collection: ED, EG, EF
Computational experiments: KM, RF
Interpretation of the results: all authors
Formal statistical analysis, visualization, and validation: KM
Writing-original draft: KM, ED, RF
Writing-editing: EG, EF, RF
Supervision: EF, RF

**Corresponding Author:**
Kyle Mahowald
mahowald@utexas.edu
University of Texas at Austin
Department of Linguistics
305 E. 23rd Street STOP B5100
Austin, TX 78712

**Abstract**

Grammatical cues are sometimes redundant with word meanings in natural language. For instance, English word order rules constrain the word order of a sentence like "The dog chewed the bone" even though the status of "dog" as subject and "bone" as object can be inferred from world knowledge and plausibility. Quantifying how often this redundancy occurs, and how the level of redundancy varies across typologically diverse languages, can shed light on the function and evolution of grammar. To that end, we performed a behavioral experiment in English and Russian and a cross-linguistic computational analysis measuring the redundancy of grammatical cues in transitive clauses extracted from corpus text. English and Russian speakers (n=484) were presented with subjects, verbs, and objects (in random order and with morphological markings removed) extracted from naturally occurring sentences and were asked to identify which noun is the subject of the action. Accuracy was high in both languages (~89% in English, ~87% in Russian). Next, we trained a neural network machine classifier on a similar task: predicting which nominal in a subject-verb-object triad is the subject. Across 30 languages from eight language families, performance was consistently high: a median accuracy of 87%, comparable to the accuracy observed in the human experiments. The conclusion is that grammatical cues such as word order are necessary to convey subjecthood and objecthood in a minority of naturally occurring transitive clauses; nevertheless, they can (a) provide an important source of redundancy and (b) are crucial for conveying intended meaning that cannot be inferred from the words alone, including descriptions of human interactions, where roles are often reversible (e.g., Ray helped Lu/Lu helped Ray), and expressing non-prototypical meanings (e.g., "The bone chewed the dog.").

## 1. Introduction

Cues like word order and morphological markings are important for conveying linguistic information (Dryer, 2002; Fenk-Oczlon & Fenk, 2008; Greenberg, 1963; Kiparsky, 1997; Koplenig et al., 2017; Levshina, 2020, 2021; Sinnemäki, 2008). But critical to theories of language are actual patterns of language use. How important are grammatical, or morpho-syntactic, cues for inferring complex meanings *in practice*? In this work, we investigate this question by exploring how redundant word order cues are for humans determining subjecthood in transitive clauses.

Simple transitive clauses, consisting of a subject (S), a verb (V), and an object (O) are commonly brought up in discussions of the importance of formal grammatical marking (e.g., how would you differentiate between "The dog bit the cat" and "The cat bit the dog"?) because they communicate fundamental linguistic information: who did what to whom. As such, they have been extensively studied across different sub-fields of language research, from linguistic theory (e.g., Comrie, 1989; Dryer, 1991) to psycholinguistics (e.g., Bates & MacWhinney, 1989; Goldin-Meadow et al., 2008), to neurolinguistics (e.g., Bates et al., 1987; Caramazza & Zurif, 1976), to computational linguistics (e.g., Palmer et al., 2013; Papadimitriou et al., 2021).

In transitive clauses, most languages use grammatical cues to differentiate grammatical roles such as subjects and objects, either through word order or through case marking and/or agreement. These rules allow different meanings to be conveyed and represented using the same set of lexical items. In English, one can use word order to differentiate "The dog bit the cat" from "The cat bit the dog." And in Russian, one can use case marking to differentiate "Dog-Nominative bit cat-Accusative" from "Dog-Accusative bit cat-Nominative".

In some utterances, though, these formal grammatical cues are not strictly necessary because lexical semantics (word meanings) strongly constrain interpretation. For example, in a sentence like "The dog chewed the bone", it is readily inferable that "dog" is the subject and "bone" is the object from the meanings of the words. In a hypothetical language in which word meanings *always* provide strong cues to interpretation, one could imagine having no constraints on word order and no case marking: "dog chew bone," "dog bone chew," "bone dog chew," "bone chew dog," "chew dog bone," and "chew bone dog" would all refer to an event of a dog chewing a bone, because alternative meanings are implausible.

If a natural language actually had the property that lexical items perfectly constrain interpretation, then it would be inefficient to have constraints like word order fixedness or grammatical marking because of the additional effort/complexity that such constraints introduce into the system.[1] Indeed, in a language that is perfectly efficient in a noiseless system (i.e., one that minimizes effort while maximizing what it can communicate; see Gibson et al., 2019 for an overview), one might imagine that the three aforementioned strategies (word order constraints, case marking, lexical semantic constraints) would neatly trade off with one another. That is, in a sentence from a language with strict word order, one might expect there to be no case information since the extra effort required to specify the case information would be extraneous. And, in a sentence like "dog chew bone", neither word order nor case information would be necessary since lexical meaning so strongly constrains the interpretation.

Researchers have long studied such tradeoffs in word order and morphological markings for conveying linguistic information (Dryer, 2002; Fenk-Oczlon & Fenk, 2008; Greenberg, 1963; Kiparsky, 1997; Koplenig et al., 2017; Levshina, 2020, 2021; Sinnemäki, 2008), and there is indeed some evidence that these cues trade off in expected ways both across languages at the typological level and within a language at the sentence level (see Levshina, 2021, for a summary and causal analysis of these factors along with a discussion of how they relate to linguistic efficiency). For instance, languages with freer word order are more likely to have case marking (Futrell et al. 2015; Greenberg, 1963; Sinnemäki, 2008), and there is behavioral evidence that this may be for reasons having to do with linguistic efficiency (Fedzechkina et al. 2016; Fedzechkina & Jaeger 2020). And, within a language, when the semantic role is more predictable for a given argument, case marking is less likely (Jäger 2007). This makes sense from a communicative perspective: if you know that "the dog" is more likely to be the subject of a sentence than "the bone", there is less pressure to mark it as a subject---a result that meshes with the presence of differential object marking across languages whereby more surprising realizations of arguments are more likely to be marked (Aissen, 2003; Jäger, 2007; Tal et al. 2022).

---

[1] Indeed, some languages have very little grammatical marking: see Ergin et al., 2018; Gil, 2013; Jackendoff & Wittenberg, 2017.

Moreover, some languages have more freedom than others in the kinds of arguments that appear in particular argument positions (e.g., whether inanimate nouns can appear in subject position). Hawkins (1986) calls this "semantic tightness"; see Levshina (2020; 2021) for quantitative measurements of this property. Whereas constraints on English arguments are relatively loose (e.g., it is not unusual for an inanimate noun to be an English subject), languages like German or Russian have much tighter constraints such that non-agentive subjects may sound unnatural (Hawkins 1986; Müller-Gotama 1994). Relatedly, in speech, there seem to be constraints on what words go early in an utterance (Stoll et al., 2009). Our work is most relevant to lexical restrictions on subjects and objects *given* verbs, rather than general restrictions based on position within an utterance.

While there is some evidence for efficient tradeoff of these factors across languages, there is also overlap in these grammatical cues. For instance, even languages where word order is sufficient to disambiguate meanings often use case marking. And, whereas a purely efficiency-based approach might suggest that a case-marked language can afford to be semantically looser, Levshina (2021) finds a correlation between semantic tightness and case marking such that case-marked languages tend to have tighter semantic tightness requirements (perhaps because semantic looseness encourages the loss of case marking).

Taken together, these findings highlight the role of **redundancy** in grammar. Grammatical cues are **redundant** when they could be removed without affecting the ability to recover the intended meaning. As redundancy is crucial for robustly transmitting information in a noisy channel (Shannon, 1948; Shannon, 1951), linguistic redundancy (Wit & Gillette, 1999; Hengeveld, & Leufkens, 2018) is a central concept in information-theoretic accounts of human language (e.g., Bentz et al. 2017; Ferrer i Cancho & Sole, 2002; Ferrer i Cancho, 2018; Jaeger, 2010; Juola, 2008; Piantadosi et al. 2012; Pimentel et al. 2021; Ehret & Szmrecsanyi, 2016; Zaslavsky et al., 2018) and has been studied previously at the level of orthographic characters (Cover & King, 1978; Shannon, 1951), sounds (Marslen-Wilson & Tyler, 1980), and words (Bentz et al., 2017), among other domains.

Why should cues like case marking and word order be redundant with meanings? One possible reason is that redundancy facilitates learning, an argument made by Tal & Arnon (2022) with artificial language learning experiments showing that an artificial language in which lexical meaning is redundant with case marking is more easily learned by naïve learners than a system without that redundancy. This is part of a larger body of work suggesting that redundancy is crucial for learning (e.g., Audring, 2014; Bates & MacWhinney 1989, Christiansen & Chater, 2016; Morgan, Meier, and Newport, 1987).

A second possibility is that, as in noisy channel models of language (e.g., Gibson et al., 2013), grammatical redundancy is crucial for inferring sentence meaning in the presence of noise (e.g., Christiansen & Monaghan 2016; Monaghan 2017).

A third possibility is simply that people sometimes want to say things that are rare or surprising or unusual. In order to do so, they need a system that lets them override lexical meanings using formal cues. That is, if an English speaker wants to say "the onion chopped the chef" (instead of the more common "the chef chopped the onion"), they can do so using word order since English word order is fixed. These possibilities are not mutually exclusive and may well all contribute to the presence of redundancy in the system.

All of these accounts assume that language users have a means of assessing the likelihood that a particular lexical item, relative to another, is a subject. That is, how do speakers come to know that "chef" is a likely subject? Past work has explored the mechanisms by which this information might be learned (such as animacy and conceptual accessibility, as described in McDonald, Bock, and Kelly, 1993; see also Chang et al. 2008; Chang, 2009 for production models using this idea). Our goal in this work is not to elucidate that mechanism but to measure the extent to which that knowledge is redundant with other kinds of information, specifically word order and case-marking.

While previous work has investigated the redundancy of grammatical cues (e.g., Levshina, 2020; Tal & Arnon, 2022; Pijpops & Zehentner, 2022), automatically estimating the information in lexical meanings of arguments is not straightforward. Levshina (2021) empirically estimated the likelihood of various lexical items being subjects or objects by estimating the mutual information between a grammatical role and a particular word. But consider a triad consisting of the verb *baffled* with the nominal arguments *map* and *traveler*. Even though *traveler* is often an agent and *map* a patient, in this case, the object is likely the inanimate *map* and it is largely unambiguous since "the map baffled the traveler" is a more likely meaning than "the traveler baffled the map." Making these kinds of inferences requires not just knowing the probability of each word appearing as a subject, but drawing on linguistic knowledge and world knowledge to ascertain how the constituent pieces fit together.

In parallel with linguistic investigations, developments in the field of natural language processing (NLP) have suggested a high level of redundancy in language, especially in word order. Much of the early success of statistical NLP was based on bag-of-words representations of sentences and paragraphs, ignoring all information about word order; nonetheless, such systems performed well on many tasks such as sentiment analysis, topic modeling, etc. (Jurafsky & Martin, 2023), suggesting that much of the information in word order is redundant from the perspective of such tasks.

Even the task of reconstructing word order from a bag of words is reasonably feasible (Gali & Venkatapathy, 2009; Horvat & Byrne, 2014): for example, Chang et al. (2008) use an incremental bag of words approach to see how often correct sentence order can be generated (across 14 typologically diverse languages) from the set of words in that sentence, relying on n-gram statistics as well as, crucially, on prominence. For all languages, they find performance far above chance (between 30% and 60%, depending on the language). As an extreme form of this result, Malkin et al. (2021) show that, by using a language model to algorithmically decide on the most probable word order for an utterance given its bag of words—effectively destroying the original order of the input words—higher performance can be achieved on downstream tasks.

These results show a high level of redundancy in aggregate statistics from the perspective of computational models; our goal is to study redundancy in a specific construction whose semantics and importance is well established, and to study redundancy for humans, who have different computations and information available to them than computational systems. To that end, the goal of the present paper is *to estimate how often grammatical cues are actually redundant with lexical meaning*, in practice, in subject-verb-object triads extracted from transitive clauses—and how that redundancy varies across typologically different languages. We propose a novel way to estimate this quantity by testing how often human participants can infer which nominal is the subject and which is the object of a transitive clause, in the absence of grammatical information and sentence context. And we develop a computational version, using artificial neural networks, of the same task. We run the model across 30 typologically diverse languages.

Specifically, we estimate the redundancy of formal cues in transitive clauses, focusing on clauses with two nominals. We presented human participants with **triads** consisting of a verb, a subject, and an object extracted from naturally occurring sentences, and asked them to guess which of the two nouns is the subject. For the human experiments, we tested native speakers of a language that relies primarily on word order cues and that has loose constraints on which nouns can appear in various argument positions (English) and a language that relies primarily on case marking and agreement cues and has relatively stricter constraints on nominal positions (Russian). The two nominals were presented in their base lemma forms, stripped of case information. We also ran a version, in both English and Russian, in which, rather than choosing which of two arguments was the subject, participants placed nouns on each side of the verb (effectively "writing" a new sentence). This version, which gave similar results, is more naturalistic than the "choose the subject" task since it does not require participants to reason about linguistic categories.

*A priori*, we expect at least some redundancy: human ability to judge which of two nominals is the subject is likely to be above chance (50%) because we know that there are some sentences (e.g., "The dog chewed the bone.") in which meaning is strongly constrained by the words. We also know that the measure of grammatical redundancy is unlikely to be 100%: for at least some sentences, grammatical cues are necessary to determine meaning ("Laura meets Petrarch." vs

"Petrarch meets Laura."). But just how redundant are sentences on average: 60%? 80%? 95%? And how consistent is that number across languages? And how will it vary between cased languages and languages without case?

Answering these questions can shed light on ongoing questions regarding redundancy and efficiency tradeoffs in grammar. If formal marking is largely distinct from information carried by the word meanings (as in the made-up example of "dog cat bite" above), then we would predict performance on this task to be closer to chance level (i.e., <60%). That is, word order and case marking would always be necessary to extract the correct propositional meaning (i.e., to determine who is doing what to whom). If, on the other hand, formal marking is fully redundant (as in the made-up example of "dog bone chew" above), then we would expect performance to be near 100%. A third possibility is that performance would differ dramatically between English and Russian, perhaps suggesting that some languages rely heavily on formal cues for encoding propositional meaning while other languages do not.

For the experiments on the computational language model, we tested a diverse sample of 30 languages spanning eight language families. The task was broadly the same as in the human experiment: we presented the model with a subject, verb, and object, and asked it to predict which of the two nominals was the subject. Because of limitations of our corpus, we used wordforms as they appeared in the corpus, not lemmas. As a result, the experiments on the language model focused on evaluating the redundancy of word order information in the presence of case marking (when it exists). This leads to a straightforward prediction that, if case-marked languages and non-cased marked languages are equally informative in terms of word order and semantic information, the extracted arguments in case-marked languages will be more easily disambiguated in our study because they have an additional information source. On the other hand, if subject/verb/object triads from case-marked languages and non-case-marked languages are equally ambiguous in this study (in which case-marking information is present), that would mean that case-marked languages, when stripped of case and word order, are *more* ambiguous than languages that never had case to begin with. In that scenario, we would conclude that case-marked languages take advantage of case-marking to convey meanings that, without case marking, would be more ambiguous based on lexical semantics alone.

In addition to shedding light on questions in typology, quantifying the level of redundancy has implications in Natural Language Processing (NLP). The current dominant paradigm in NLP involves training neural network models on huge amounts of data on a word prediction task (Devlin et al., 2019; Brown et al., 2020). Such models seem to learn sophisticated syntactic and semantic machinery, as evidenced by model analyses (e.g., Hewitt & Manning, 2019; Linzen and Baroni, 2021; Tenney et al., 2019) and strong task performance on linguistically challenging tasks (e.g., Linzen et al. 2016; Gulordava et al., 2019; Srivastava, 2022; Finalyson et al., 2021). But a recent body of work suggests that these models are effective on a variety of syntactic and semantic tasks even when they are trained on word-order-scrambled input or without access to word order information – showing drops of just a few percentage points when compared to models trained on regular text (Abdou, 2022; Papadimitriou et al., 2022; Clouatre et al., 2021; Hessel & Schofield, 2021; Malkin et al. 2021; Ravishankar et al., 2021; Sinha et al., 2021). If it turns out that word order information is often redundant with word meanings (over, say, 90% of the time), then these findings may be unsurprising: just the presence of the lexical items alone would be enough to recover the meaning *most of the time*. So, on average, performance would seem to drop, even as the model was failing on sentences that really did depend on sensitivity to grammatical cues. But if, say, grammatical cues were redundant only 60% of the time, these results would be more puzzling. Thus, we believe there is value to the NLP community in measuring just how redundant these cues are.

To foreshadow our results, we found that for a large majority of sentences across typologically diverse languages, humans and a computational language model can correctly infer the subject of a transitive sentence, without word order information, for about 85%-90% of sampled sentences. However, doing so is not possible for ~10-15% of sentences. In our computational study, numbers were similar to the human experiments and held across a variety of languages (with the exception of Chinese, where performance was lower, likely due to issues with how the model was constructed). As we discuss in more depth in the computational modeling section, note that the computational study allowed models access to morphological information and so the estimates here have slightly different interpretations depending on the degree of marking in that language.

It is worth noting that, in one sense, our performance estimates are conservative in that participants (humans or a computational model) have access to *less* information than is generally available during language comprehension. Participants see only the triad of subject, verb, and object, and not any other arguments of the verb, modifiers of the nominals, or any other parts of the sentence, or the preceding context. Moreover, across many languages (including English and Russian), we would expect transitive sentences with a pronoun subject or object (which we exclude from our study, but which make up most transitive sentences cross-linguistically; Ariel, 1991; Du Bois, 1987) to be nearly perfectly classifiable on this task. Because these other sources of information can disambiguate the argument structure, the accuracies that we report are best interpreted as approximate lower bounds on accuracy, given only the information directly present in the lexical semantics of the verb, its subject, and its object.

At the same time, our estimates are based only on classifying the subject, verb, and object of transitive clauses. It would of course be more difficult for both our human participants and computational models to infer the relationship between all the elements of a long sentence (e.g., not just subjects, verbs, and objects but prepositional phrases, relationships between subclauses, etc.), as attempted in computational work (e.g., Chang et al., 2008, Gali & Venkatapathy; Horvat & Byrne, 2014; Malkin et al., 2021). Thus, these measures of redundancy should not be taken as measures of grammatical redundancy *in general* but of redundancy in SVO triads extracted from transitive clauses.

## 2. Human experiments

### 2.1 Experiment 1: English

We conducted eight experiments (Experiments 1a-f in English, and Experiment 2a-b in Russian) where extracted examples of transitive verbs with subjects and objects were presented in a scrambled order and with morphological markers removed. If human participants can guess which noun is the subject (or correctly place the subject in Experiments 1e and 2b)t, that would indicate that the subject-object distinction can be recovered based on the meanings of the nouns and the verb alone, leaving formal marking redundant.

We extracted clauses containing transitive verbs from parsed corpora and reduced each such clause to a subject-verb-object (SVO) triad: the head noun of the subject noun phrase, the head noun of the object noun phrase, and the head lexical verb, each converted to a suitable form to remove morphological marking such as case and agreement which could be used to recover which noun is the subject. Therefore, when an SVO triad was presented in a shuffled order, it contained neither word order nor morphological cues to propositional meaning. On each trial, participants saw a verb that was followed by two nouns (whether the subject or the object appeared first on each trial was random) and were asked to choose one noun, which they think is the subject, or do-er, of the action described by the verb.

We ran 4 versions of the experiment in English. A similar set of materials was used across the four experiments; Experiments 1b-d were performed to ensure the robustness and replicability of the results obtained in Experiment 1a.

**Methods: Experiment 1**

*Participants*

Across four experiments, we recruited 395 participants on Amazon Mechanical Turk: 100 in Experiment 1a with 21 excluded for not being native speakers or performing below chance (in Experiments 1b-d, we used catch trials to detect guessing, as detailed below, and excluded participants who answered fewer than 75% of catch trials correctly); 100 in Experiment 1b with 19 excluded; 100 in Experiment 1c with 16 excluded; and 95 in Experiment 1d with 10 excluded. The exclusions left 329 participants for analysis (79 in Experiment 1a, 81 in Experiment 1b, 84 in Experiment 1c, and 85 in Experiment 1d), comprising 309 unique participants (some appeared in multiple experiments; their inclusion does not qualitatively affect the results). The experiment took approximately 20 minutes to complete, and participants were compensated $3.00 for their time.

*Experimental materials*

We extracted English sentences from the Universal Dependencies English Web Treebank (EWT). A triad was identified as any verb (with universal part-of-speech tag *VERB*) with exactly one dependent of type 'subject' (*nsubj*) and exactly one dependent of type 'object' (*obj*). Triads where the subject, the object, or both were pronouns (n=3,655) were excluded because pronouns contain case-marking information. Of the triads with either OSV or SOV word order, 7 were mis-parsed (e.g., contained a verb in the object position), and were consequently flipped (e.g., "remedies the trustee is seeking" "the trustee is seeking remedies") to constitute an SVO triad using information from the rest of the sentence. While there has been a large amount of work on what constitutes subjecthood (e.g,. Dixon, 1994; Comrie, 1989; Keenan, 1976; Tollan, 2019, etc.), we use the Universal Dependencies notion of subjecthod, which seeks to pick out the "syntactic subject and the proto-agent of a clause" (see Nivre et al., 2016, for further discussion of these annotations).

This initial filtering left 631 triads (14.7% of the original set; transitive sentences with two full nominal arguments are generally rare cross-linguistically; Ariel, 1991; Du Bois, 1987). Further, 42 triads were excluded for various reasons (e.g., offensive content or repeats), leaving 589 triads, and 278 of these were slightly edited. In particular, in 136 triads, the verb's tense was changed to past simple; in 48 triads, the verb phrase was corrected to ensure that the intended meaning is conveyed (e.g., *threw -> threw up*); in 101 triads, the agent or the patient noun phrase was corrected to ensure that the intended meaning is conveyed (e.g., *Rita -> Hurricane Rita*); finally, in 30 triads, possessive pronouns modifying the agent or the patient were deleted because they could provide cues to the dependency structure. For Experiment 1a, the 589 triads were distributed across 5 experimental lists (118 triads in Lists 1-4 and 117 triads in List 5) for presentation. (For this and all other experiments, the materials, including the original, excluded, and edited triads, are available at OSF: https://osf.io/kbtga/.) For Experiment 1b, we additionally excluded 20 and edited 330 triads, and distributed the remaining 569 triads across 5 experimental lists (114 triads in Lists 1-4 and 113 triads in List 5). For Experiment 1c, we additionally excluded 50 triads, and distributed the remaining 519 triads across 5 experimental lists (104 triads for Lists 1-4, and 103 triads for List 5). Finally, for Experiment 1d, we randomly sampled 500 triads from the set of 519, and distributed them across 5 experimental lists (100 each).

In Experiments 1b-d, 20 triads with clear grammatical roles (animate subjects, inanimate subjects, and a prototypical subject-object relationship with respect to the verb: e.g., *pharmacist prescribed medicine*) were included in each list as 'catch trials' to ensure that participants engage with the task. Catch trials were randomly interspersed with the critical triads and were excluded from the critical analyses. Participants who did not identify the subject correctly in 15 or more of the catch trials were excluded.

*Procedure*

At the beginning of the task, participants were provided with an example trial that was not a part of the experimental stimulus set (*chewed bone dog*) and told that the correct answer is *dog* because dogs chew bones. All trials were presented on one web page (the order was randomized for each participant) with brief instructions (i.e., *Click on the do-er of the action*) appearing above each triad as a reminder. Prior to the critical task, participants were asked to indicate their native language and told that the payment is not contingent on their answer.

In Experiments 1b-d, the instructions were edited to include a description of what nouns and verbs are (i.e., nouns - words that denote people, things, phenomena, and verbs - words that denote actions), and participants were asked to guess who is doing the action described by the verb (because based on informal feedback and the presence of some participants with below-chance performance in Experiment 1a, the term "agent/do-er" appeared to be confusing for some participants). Experiment 1e avoids this terminological confusion entirely by asking participants to place nouns on either side of the verb.

**Results**

*Overall performance*

The results were similar across the four human experiments (Figure 1, top panel). In ***Experiment 1a***, the mean percent correct, across participants, was 88.9 [95% CI on participant means 87.8%, 89.9%]. The item with the maximum accuracy had correct answers 100% of the time, the item with the lowest accuracy was correct 0% of the time, and the median item accuracy was 100%. 80.5% of items had over 80% accuracy, and 71.1% had over 90% accuracy.

In ***Experiment 1b***, one item was excluded from the analysis because the participants reported a display error. The mean percent correct, across participants, was 88% [95% CI on participant means 86.8%, 89.2%]. The item with the maximum accuracy had correct answers 100% of the time, the item with the lowest accuracy was correct 0% of the time, and the median item accuracy was 100%. 79.1% of items had over 80% accuracy, and 71.2% had over 90% accuracy.

In ***Experiment 1c***, the mean percent correct, across participants, was 89.7 [95% CI on participant means 88.8%, 90.6%]. The item with the maximum accuracy had correct answers 100% of the time, the item with the lowest accuracy was correct 5.9% of the time, and the median item accuracy was 100%. 82.5% of items had over 80% accuracy, and 71% had over 90% accuracy.

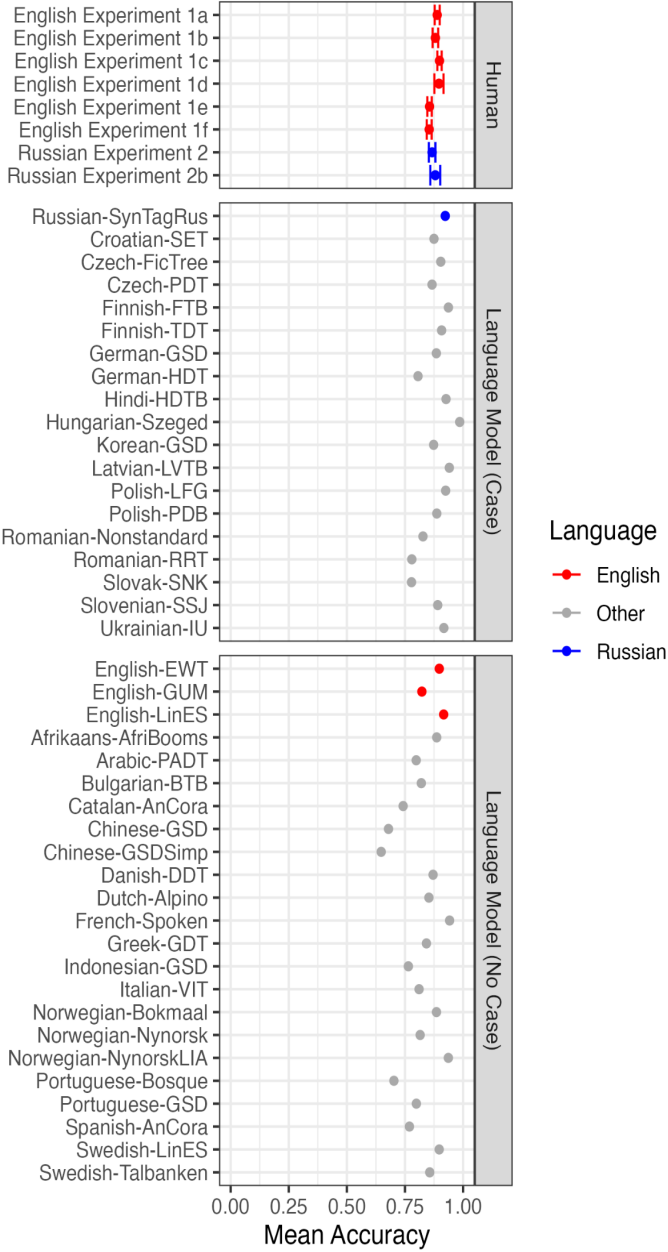

**Figure 1.** The x-axis shows the mean accuracy for each experiment. The top panel shows human experiments (for English and Russian); performance is shown with 95% confidence intervals. The two bottom panels show computational experiments on the Universal Dependency corpora across languages, split into languages that use case (middle panel) and languages that do not (bottom panel). Performance is consistently high and comparable between human participants and the language model. (Note that we show just the noun portion for Experiment 1e in this figure, to make the points comparable.)

Finally, in ***Experiment 1d***, the mean percent correct, across participants, was 89.6% [95% CI on participant means 87.7%, 91.6%]. The item with the maximum accuracy had correct answers 100% of the time, the item with the lowest accuracy was correct 5.6% of the time, and the median item accuracy was 94.7%. 84.6% of items had over 80% accuracy, and 68% had over 90% accuracy. Across experiments, only for 5% of the items was accuracy lower than 50%, suggesting that most items in the sample could be guessed at a level better than chance.

These results suggest that lexical-semantic information (word meanings) alone is sufficient to identify the subject of a transitive verb in approximately 89% of the cases.

*Animacy analysis*

To better understand this trend and given that animacy is a strong cue to subjecthood in language (Ariel, 1991; Comrie, 1989; Dahl, 2008; Dixon, 1994; Everett, 2009; Osgood, 2013), we categorized each subject and object across the entire set of materials used in Experiments 1a-d as animate or inanimate (collapsing across experiments). Each item was coded by 2 authors (E.D. and K.M.). The disagreement rate was low (31 out of 1090 items) and disagreements were resolved through discussion.

We then ran a post-hoc analysis exploring the accuracy across the 4 'conditions': animate subjects + animate objects (n=88 triads; e.g., *Petrarch meets Laura*, *Johnson deployed troops*), animate subjects + inanimate objects (n=436; e.g., *guys cooked food*), inanimate subjects + animate objects (n=48; e.g., *shops have owners*), and inanimate subjects + inanimate objects (n=518; e.g., *alternatives do not have requirements*).[2]

Triads with animate subjects and inanimate objects were the overall easiest to classify, as can be seen in Figure 2. The inverse triads (with inanimate subjects with animate objects) were the most difficult to classify and generated a high rate of incorrect guesses. The two symmetric conditions, where both the subject and object are animate or both are inanimate, fell in between, but the both-animate triads were harder. This is likely because these sentences tend to be semantically "reversible" (Caramazza & Zurif, 1976): "Petrarch meets Laura" is just as plausible as "Laura meets Petrarch" (but not always: e.g., "Johnson deployed troops"). The both-inanimate triads tend to be less reversible and thus might offer clearer cues (e.g., *camera requires reboot*, compared to the less plausible *reboot requires camera*).

To assess the statistical significance of these animacy-related differences, we ran a mixed effect logistic regression predicting whether the answer was correct based on the animacy of the subject, the animacy of the object, and their interaction. Following Barr et al. (2013), we started with a maximal random effect structure but found that it did not converge. We iteratively removed elements of the random effect structure (removing first the correlation parameters and then the lowest variance elements). We were left with a random intercept term for item and random intercepts for participants, along with by-participant slopes for subject animacy and object animacy. The fixed effect term for subject animacy was positive (β=.60) and negative (β=-1.63) for object animacy. As predicted, these results suggests that participants were more likely to be correct when the subject was animate and more likely to be correct when the object was inanimate.

---

[2] Animacy is perhaps the strongest cue to subjecthood, but there are others attested in the literature (e.g., discourse status, information structure, imageability, accessibility). We leave the exploration of these to future work.

To assess the significance of these terms, we ran a likelihood ratio test comparing the full model to a model with no fixed effect predictors for animacy. The full model provided a significantly better fit ($\chi^2(3) = 139.44$, $p < .00001$), suggesting that animacy information is a useful predictor as to whether participants can successfully guess the subject of the sentence. We also ran likelihood ratio tests comparing the full model to models with identical fixed effect structures but without the effect of subject animacy, object animacy, and the interaction, respectively. While the effects of subject animacy ($\chi^2(1) =14.58$, $p < .001$) and object animacy ($\chi^2(1) =98.0$, $p < .00001$) were significant, the interaction was not ($\chi^2(1) =2.39$, $p=.12$).

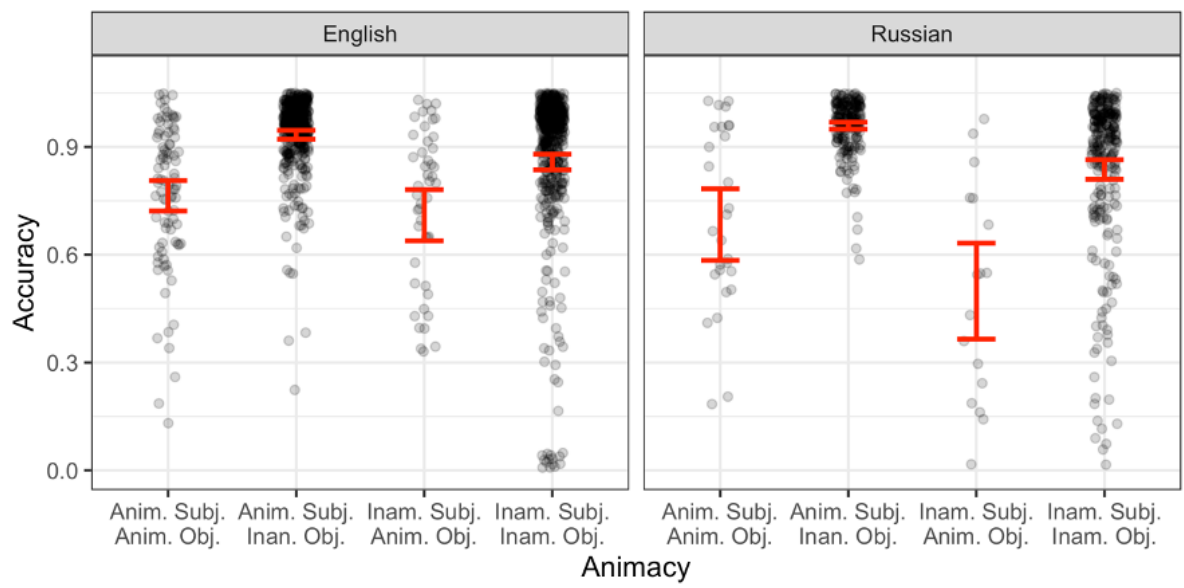

**Figure 2.** Accuracy as a function of animacy, for English and Russian human participants. The individual data points represent means for individual sentences. Error bars represent 95% confidence intervals over sentences. Because there were fewer sentences overall in Russian, the conditions with fewer naturally occurring examples (animate subject + animate object, inanimate subject + animate object) are particularly noisy, as is reflected by the large error bars. In both English and Russian, sentences with animate subjects and inanimate objects exhibited the highest accuracies, and sentences with inanimate subjects and animate objects exhibited the lowest accuracies.

Note that, although the interaction term was not significant, the effect of object animacy is more than twice as large as the effect of subject animacy. This asymmetry is consistent with the observation that, across languages, differential object marking—the use of optional morphological marking on objects, often on animate objects rather than inanimate objects—is more common than differential subject marking (Aissen, 2003; Haspelmath, 2019; see Section 5.2 for more discussion of connections to differential object marking).

**Interim Discussion**

Based on Experiments 1a-d, we obtained an estimate of the redundancy of word order in simple transitive clauses. But in these experiments, 85.3% of transitive sentences in our initial sample were excluded because they contained pronouns (an estimate broadly consistent with cross-linguistic findings as to the rarity of transitive sentences with multiple full nominal arguments; Du Bois et al., 2003). These omitted materials often contained grammatical information, because many English pronouns are marked for case. Because of these exclusions, our estimate of human performance on the task (~88%) reflects the redundancy of word order a) on nouns and b) in the absence of case marking since there is no case marking on English nouns.

Moreover, there is a limitation to the method in that we ask participants to select the agent or do-er of the action, which can be confounded with thematic role. In the next set of experiments, we seek to address both of these issues by including pronouns and devising a variant of the original task in which participants place the arguments on either side of the verb.

Specifically, in Experiment 1e, we include English pronouns (with case information when it's there, as in words like *I* and *me*) and explicitly test how their redundancy contributes to the overall estimate. In Experiment 1f, we include English pronouns but remove case information by having case-restricted words appear in both possible forms (e.g., *I/me*, *he/him*).

**2.2 Experiment 1e-f: English Pronouns**

In the first set of experiments, participants were instructed to select the agent or do-er of the action denoted by the verb. As a result, it is possible that the human experiments did not capture the judgment of the true grammatical roles, unlike the computational experiments, where the models were trained to identify the subject of the sentence. To address this issue, we conducted two additional human experiments where participants generated sentences using the words from the triad, by placing the specified nouns on either side of the specified verb.

We also included pronouns in half the sentences in these experiments, in order to explicitly evaluate pronouns rather than relying purely on the back-of-the-envelope calculations above. In Experiment 1e, we did not strip case information from pronouns. We also ran Experiment 1f, which was identical to Experiment 1e except that we presented the pronouns in a case-neutral way such that, if the pronoun was "I" it appeared as "I/me" and if it was "him" appeared as "he/him."

**Methods: Experiment 1e**

*Participants*

We recruited 101 participants on Prolific. Two participants were excluded because their data were not recorded properly. One additional participant was excluded for reporting a limited level of proficiency in English, leaving 98 participants for the analysis. The experiment took on average 20 minutes to complete, and participants were compensated at a rate of $12.00/hour.

*Materials*

For the noun condition, we randomly sampled 50 triads from the set of 519 triads used in Experiments 1a-d. For the pronoun condition, we randomly sampled additional 50 triads from the set of previously excluded triads and slightly edited 36 of them. In particular, in 20 triads, the verb's tense was changed to the past tense; in 3 triads, the verb phrase was corrected to ensure that the intended meaning is conveyed; in 4 triads, the agent or the patient noun phrase was corrected to ensure that the intended meaning is conveyed; finally, in 23 triads, determiners were deleted. Each triad in the pronoun condition contained only one pronoun either in the subject or object position.

*Procedure*

Participants were presented with three words on the screen: nouns and pronouns were at the top of the screen, each on a separate line, and a verb was in the middle of the page and had one blank space to the left and another -- to the right of it. Participants were instructed to place one word to the left of the verb and the other to the right to make a sentence that these words could have been taken from. They were also told that they may change the form of the words to ensure that the sentence was grammatically correct. Participants were provided with an example trial that was not a part of the experimental stimulus set (*played balloon child*) and told that a possible sentence is "*A child played with the balloon*". Each trial was presented on a separate web page (the order was randomized for each participant) with the brief instructions (i.e., *Please use 2 words above to make a simple sentence with the verb below*) appearing above each triad as a reminder. At the end of the task, participants were asked to indicate their level of proficiency in English and told that the payment is not contingent on their answer.

**Results**

The mean percent correct, across participants, was 88.91% [95% CI on participant means 88.24%, 89.57%]. The item with the maximum accuracy had correct answers 100% of the time, the item with the lowest accuracy was correct 3% of the time, and the median item accuracy was 96.94%. 80% of items had over 80% accuracy, and 75% had over 90% accuracy. Critically, the mean accuracy for the noun condition was 85.57% [95% CI on participant means 84.62%, 86.52%] and the mean accuracy for the pronoun condition was 92.24% [95% CI on participant means 91.51%, 92.98%].

When comparing just the noun conditions (excluding pronouns), this method gives similar estimates (86% compared to 89%) to the method used in Experiments 1a-d.

**Methods: Experiment 1f**

*Participants*
We recruited 100 participants on Prolific. Two participants were excluded because their data were not recorded properly. Four additional participants were excluded for reporting a limited level of proficiency in English, and one participant was excluded for not providing any responses, leaving 93 participants for the analysis.

*Materials*

Materials were the same as in Experiment 1f except that any pronoun which contained case information (e.g., *I*, *me*, *him*, *her*, etc.) appeared with an alternative (e.g., *I/me* or *he/him*).

*Procedure*

The procedure was the same as for Experiment 1e.

**Results**

The mean percent correct, across participants, was 85.87% [95% CI on participant means 83.96%, 85.78%]. The item with the maximum accuracy had correct answers 100% of the time, the item with the lowest accuracy was correct 0% of the time, and the median item accuracy was 93.55%. 75% of items had over 80% accuracy, and 58% had over 90% accuracy. The mean accuracy for the noun condition was 85.41% [95% CI on participant means 84.36%, 86.47%; compared to 85.57% for the nouns in Experiment 1e] and the mean accuracy for the pronoun condition was 84.32% [95% CI on participant means 83.24%, 85.40%; compared to 92.24% for the pronouns in Experiment 1e].

Note that, while the noun conditions between Experiments 1e and 1f are extremely similar, the pronouns differ in a predictable way (decreasing markedly in Experiment 1f) since case information is available for English pronouns in Experiment 1e but not in Experiment 1f. This suggests that, when case information is controlled for, pronouns and nouns behave similarly in our experiment.

**Interim Conclusion**

Across two different methodologies, we found similar estimates of redundancy, 85%-90%, for sentences extracted from an English language corpus. The redundancy was even higher (more like 92%) for sentences involving pronouns when those pronouns included case information, but crucially not when case information was stripped (dropping the estimate back to 84%). These results are informative as to the redundancy of grammatical cues in transitive clauses for a language with no case marking: English. In the next section, we consider a typologically different language, Russian, which has case marking.

**2.3 Experiment 2: Russian**

The goal of Experiment 2 was to investigate the same question as in Experiments 1a-1d in a typologically distinct language. We chose Russian because, unlike English, Russian word order is highly flexible whereas most words are morphologically marked with case and/or agreement.

**Methods: Experiment 2a**

*Participants*

We recruited 89 participants (a mix of Russian native speakers residing in the US and those residing in Russia) through word of mouth. 10 were excluded for answering fewer than 75% of catch trials correctly, leaving 79 participants for analysis.

*Experimental materials*

1,047 SVO triads were extracted from the SynTagRus corpus from Universal Dependencies. A similar procedure was used to the one used for English to identify transitive clauses and to extract the triads. Triads where the subject, the object, or both were pronouns (n=218) were excluded because pronouns contain case marking information. Further, 226 triads were excluded for various reasons (e.g., mis-parsing, or containing fixed expressions, which would facilitate the identification of the subject), leaving 603 triads (57.5% of the original set), and 601 of these (99.7%) were slightly edited. In particular, in 601 triads, the verb was changed to the infinitive form; in 238 triads, the agent or the patient was corrected to the nominative case; in 73 triads, the agent or the patient noun phrase was corrected to ensure that the intended meaning is conveyed; finally, in 50 triads, possessive pronouns modifying the agent or the patient were deleted because they could provide cues to the dependency structure. (The original, excluded, and edited triads are available at OSF: https://osf.io/kbtga/.)

We randomly sampled 500 triads from the set of 603, and distributed them across the 5 experimental lists (100 each). Additionally, as in Experiments 1b-d, 20 triads with clear grammatical roles (animate subjects, inanimate objects, and a prototypical subject-object relationship with respect to the verb; e.g., *родители купить подарки – parents buy gifts*) were included in each list as 'catch trials' to ensure that participants engage with the task. Catch trials were randomly interspersed with the critical triads and were excluded from the critical analyses.

*Procedure*

The procedure was identical to that used in Experiment 1a-d, except that participants were not recruited through Amazon Mechanical Turk, but were provided with a link for the task.

**Results**

The mean percent correct, across participants, was 86.7 [95% CI on participant means 85.3%, 88.1%]. The item with the maximum accuracy had correct answers 100% of the time, the item with the lowest accuracy was correct 0% of the time, and the median item accuracy was 94.4%. 79% of items had over 80% accuracy, and 65.6% had over 90% accuracy.

To explore the effects of animacy, similar to what we did for English, we categorized each subject and object as animate or inanimate, and explored the accuracy across the 4 'conditions': animate subjects + animate objects (n=29 triads), animate subjects + inanimate objects (n=257), inanimate subjects + animate objects (n=197), and inanimate subjects + inanimate objects (n=27). Animacy was coded by a native Russian speaker (E.D.) and points of uncertainty were discussed with E.F. Similar to what we found for English, triads with animate subjects and inanimate objects were the overall easiest to classify, as can be seen in Figure 2. The inverse triads (with inanimate subjects with animate objects) were the most difficult to classify.

The mixed effect models (same as those described for the English data, but with a full random effect structure except for the correlation parameter) showed that, as with English, animate subjects were more likely to be identified correctly ($\beta = 1.64$; $\chi^2(1) = 26.76$, $p < 0.0001$ by a likelihood ratio test comparing the full model to a model without the subject-animacy fixed effect), and animate objects were *less* likely to be identified correctly ($\beta = -2.90$; $\chi^2(1) = 76.385$, $p < 0.001$). As with English, the interaction term was not significant ($\beta = -.53$, $\chi^2(1) = .40$).

**Redundancy of word order in Russian when case information is available**

Because pronouns were excluded and case information was stripped from nouns before running the experiment on Russian triads, our results reflect the redundancy of word order and case information *combined*. We could also ask about the redundancy of word order information alone, by studying sentences where case marking is present and pronouns are not excluded. To explore that, we analyzed triads from 200 transitive sentences from our initial corpus sample, without removing case marking and including sentences with pronouns. After excluding 6 misparsed sentences, we were left with 194 triads for analysis. Of these, 168 (86.6%) triads were unambiguous based on morphological marking. The remaining 13.4% were similar to the sentences that we used in Experiment 1 in that they were not disambiguated by case. Assuming that this 13.4% can be guessed at a similar rate as in our sample of full NPs (86.7%), then we arrive at an overall estimate for sentences that include pronouns and case information on nouns: $(1 * .866) + (.867 * .134) = \sim 98\%$. Compared to our estimate of information in word meanings alone (86.7%), this estimate confirms the observation that word order cues are often redundant in Russian (~98% of the time).

**Comparison of Russian data to English data**

To formally assess whether the Russian data pattern differed significantly from the English data pattern, we fit a mixed effect model predicting whether the answer was correct, based on language (English or Russian), with random intercepts for subjects and items (slopes prevented convergence). The accuracies in the Russian experiment were slightly lower but not significantly so ($\beta = 0.25$, $\chi^2(1) = 2.34$, p =. 13).

We also compared the animacy analyses in English and Russian by running a mixed effect logistic regression predicting whether a particular noun was a subject (as opposed to an object) based on the animacy status of the subject and its interaction with the language (English vs. Russian). Following Barr et al. (2013), we fit a maximal random effect structure with random intercepts and slopes for subjects and items. We removed the correlation term for convergence. Russian sentences had a significantly stronger relationship with animacy ($\beta = 1.01$, $\chi^2(1) = 14.53$, $p < .01$). We did the same for the animacy status of the object (but still predicting subject) and also found a significant interaction such that in Russian, the animacy of the object was a stronger predictor of the status of the answer ($\beta = -.95$, $\chi^2(1) = 6.76$, $p < .01$). These results offer additional evidence that Russian is semantically tighter than English since it is clear that the animacy of the nouns is a stronger cue.

## 2.4 Experiment 2b

The goal of Experiment 2b was to investigate the same question as in Experiment 1e in a typologically distinct language.

**Methods: Experiment 2b**

*Participants*

We recruited 100 participants on Prolific (a mix of Russian native speakers residing in the US and those residing abroad). Five participants were excluded because they indicated their level of proficiency in Russian as "basic", two additional participants were excluded because they failed to perform the task (i.e., typed in meaningless strings of letters) leaving 93 participants for the analysis. The experiment took on average 20 minutes to complete, and participants were compensated at a rate of $12.00/hour.

*Materials*

We sampled 50 triads (consisting of full noun arguments) from the set of 603 triads used in Experiment 2a such that at least one noun in the triad was not nominative-accusative syncretic. After we collected the data, we noticed that 1 triad took genitive and not accusative case, and it was excluded from the analysis.

*Procedure*

The procedure was identical to that used in Experiment 1e, except that participants were additionally allowed to change the form of the verb to ensure that the sentence was grammatically correct. At the end of the task, participants were asked to indicate their level of proficiency in Russian and told that the payment is not contingent on their answer.

**Results**

The trials where participants did not use at least one target word from the triad or inflected one of the nouns in a case other than nominative or accusative were coded as NA and were excluded from the subsequent analysis. The trials where the subject was inflected in the nominative case and the object was inflected in the accusative case were coded as correct regardless of the word order (e.g., *металлург (nom) сменил академика (acc) / академика (acc) сменил металлург (nom) / a metallurgist replaced an academic*. The rest were coded as incorrect.

The mean percent correct, across participants, was 88.03 [95% CI on participant means 85.90%, 90.17%]. The item with the maximum accuracy had correct answers 100% of the time, the item with the lowest accuracy was correct 31% of the time, and the median item accuracy was 94%. 81% of items had over 80% accuracy, and 75% had over 90% accuracy.

**Interim Conclusion**

Our results from human participants reveal two striking patterns. First, in the majority of instances in usage, formal marking of the subject-object distinction is redundant: the subject of a transitive clause can be identified from the lexical semantics of the nouns and the verb alone, without any need for marking via word order, case, or agreement. And second, the accuracy with which people can identify the subject of a transitive clause given this information is the same (~85-90%) in two distinct languages, English and Russian. The similarity of these accuracy scores is all the more surprising considering the differences between these languages (English predominantly relying on word order cues, and Russian – on case marking and agreement), between the participant pools, and between the materials—the English triads and Russian triads were not translation-equivalent; they were drawn from independent corpora.

In the next section, we seek to expand this work to include a wider variety of languages. To do so, we turn to a computational experiment.

### 3. Experiment 3: Computational experiment

To evaluate the broader cross-linguistic generality of these patterns, we carried out a number of computational experiments using 42 Universal Dependencies 2.5 treebanks of 30 languages across eight language families, in which we study the extent to which the subject of a triad can be identified based on *word embeddings*—representations of the meaning of a word in terms of high-dimensional vectors, which have rapidly become the state-of-the-art method for representing word meanings in the field of natural language processing (Devlin et al., 2019; Mikolov et al., 2013; Pennington et al., 2014). Although these embeddings are engineering artifacts, they capture human semantic judgments on diverse tasks (e.g., Pereira et al., 2016).

In particular, we report a series of computational experiments that examine the extent to which word order is redundant as a cue to the subject-object distinction. Specifically, inputting two nouns and a verb, we train a neural network model to predict, based on word embeddings, which of the two nouns is the subject of the sentence and which is the object.

Optimally, we would also study the information content of case as a grammatical cue, and of word order in the absence of case, by comparing embeddings with and without case information. However, due to corpus limitations, we are not able to use or derive embeddings without case information. As a result, for languages that use case marking, model accuracies reflect contributions from both lexical semantics and case marking; for languages that do not use case marking, model accuracies more veridically reflect contributions from lexical semantics alone.

**Methods: Computational Experiment**

*Corpus extraction*

Similar to what we did for the human experiments, SVO triads were extracted from the Universal Dependencies 2.5 corpora by searching for all verbs with exactly one dependent of type 'subject' (*nsubj*) and exactly one dependent of type 'object' (*obj*). We included only languages for which we could extract at least 1,600 triads by these criteria.

*Word embeddings*

Our goal was to determine the extent to which the subject of an SVO triad could be identified solely based on the word meanings. To do so, we represented each distinct word as an ***embedding***: here, a point in a 300-dimensional space. Specifically, we used fastText, a set of word vectors constructed by training on the Wikipedias of a large number of languages (Bojanowski et al., 2017). Because fastText does not provide vectors for lemmas, only for wordforms, it was not possible to eliminate morphological information as we did in the human experiments. To get a vector representation of a triad as a whole, we concatenated these vectors (first the verb, then the subject, and the object, the latter two in a random order) to form a 900-dimensional vector.

*Classifiers*

Once we represented these triads as vectors, we fit classifiers to predict subjecthood (whether the first noun in the shuffled SVO triad is the subject or the object). Following standard practice in natural language processing, we used feedforward neural networks as classifiers. The neural network takes the 900-dimensional triad vector as input, then runs it through two layers of hidden units with ReLU activation (Nair & Hinton, 2010), with softmax activation for the final output. The number of hidden units is determined on a per-language basis by hyperparameter search, as described below.

*Training and validation*

We trained neural network subjecthood classifiers by backpropagation using the Adam optimizer (Kingma & Ba, 2014). For each corpus, we fit several neural network classifiers, with the learning rate drawn from {.001, 0001}, and with the number of hidden units in the first layer drawn from {32, 64, 128}, and the number of hidden units in the second layer drawn from {32, 64, 128}, a total of 18 classifiers per corpus. Each Universal Dependencies corpus has separate training, development, and test sets defined by the Universal Dependencies project. Individual classifiers were trained on the UD training set. For each corpus, we selected the best-performing classifier by taking the classifier with the highest accuracy on the UD development set. (The Universal Dependencies datasets come with predefined train-dev-test splits consisting of 80%-10%-10% of the data; which were used here.) We analyzed the final results based on accuracy on the UD test set. This procedure of holding out data guards against overfitting: final accuracy is always evaluated based on data that was not used during the process of fitting or optimizing the classifier.

**Results**

Test-set classifier accuracies are shown in the middle panel (for languages with case marking) and the bottom panel (for languages without case marking) of Figure 1. All classifiers performed better than chance on the test set. The median accuracy of the classifiers across corpora was 87% [mean of 85% with a 95% CI 83%, 88%], with a minimum of 65% for the simplified Chinese GSD corpus and 68% for the standard Chinese GSD corpus and a maximum of 99% for the Hungarian-Szeged corpus. Half of the corpora fell between 81% and 91% in accuracy. The three English corpora in the sample fell between 82% and 91% accuracy, and the Russian corpus had 92% accuracy.

**Discussion**

These results were similar in magnitude to those from the human experiments. There was some variation across languages, but there was also variation between different corpora from the same language (e.g., accuracy was 90% for English EWT, a corpus of web text, but only 85% for English GUM, a corpus of mixed genres). Some of the anomalously low accuracies may be due to issues with the word embeddings—for example, the Chinese corpora have low accuracies, possibly because the fastText vectors use embeddings of character sequences, and this scheme may be less well suited to Chinese characters than to Latin characters. More generally, these computational estimates can be thought of as lower bounds on the potential accuracy of this task since better architectures and larger data sets could well lead to improved performance.

Because, as described in Methods, the study used wordforms and not lemmas, languages with case marking have more information available to the model than languages without case marking and than our human experiments in English and Russian (where we excluded case information). Languages without formal case marking have, in principle, the same information available as our human experiments. We categorized whether or not a language was case-marked by assessing whether it has direct morphological marking on its subject and direct object (meaning Spanish, which does mark indirect objects, is not considered a cased language). As expected, case-marked languages exhibited better performance (89% on average) than languages without case marking (82% on average). To assess the statistical significance of this difference, we ran a mixed effect model predicting the mean accuracy for a particular corpus based on a binary coded variable for whether the language has case marking, with a random intercept for language. Including the case-marking variable significantly improved fit by a likelihood ratio test comparing the full model to a simpler model without the case-marking predictor ($\beta=.07$, $\chi^2(1) = 7.52$, $p<.01$). Crucially though, even for languages with no case marking, performance was well above chance suggesting that word meanings alone are enough for the model to differentiate the subject and object.

## 4. General Discussion

In this study, we used a combination of human experiments and experiments with a computational language model to evaluate how often the correct propositional meaning of transitive clauses can be inferred from just the meanings of the keywords in the absence of formal grammatical cues, like word order and case and agreement markers. Across typologically diverse languages, we found that for the majority of sentences, formal cues were redundant, although case markers did show a small contribution in the experiments with the computational language model such that the model was better able to identify the subject in case-marked languages than in languages without case-marking. For human participants, animacy was an important cue to subjecthood (see also Ariel, 1991; Comrie, 1989; Dahl, 2008; Dixon, 1994; Everett, 2009; Osgood, 2013).

It is commonly argued that different formal grammatical cues trade off in conveying meaning efficiently. For example, word order might trade off with the use of morphology (e.g., Fenk-Oczlon & Fenk, 2008; Koplenig et al., 2017; Levshina, 2020, 2021; McFadden, 2003). In the case of transitive clauses, if the subject of a verb is distinguished by morphology, then there should be no need to mark it by word order, and vice versa. But this reasoning presupposes the general utility of formal cues for conveying complex meanings.

Contra this presupposition, we showed that i) in English and Russian, both word order and morphology are largely redundant with the information conveyed by word meanings; and ii) across a variety of languages in our computational sample, word order is largely redundant. This redundancy is present even for languages that lack case-marking systems. And although the language model performs better on case-marked than non-case-marked languages, this difference is relatively small (a 7% difference in accuracy, on average) and we observe that some models trained on non-case-marked languages actually outperform models trained on case-marked languages (e.g., English-EWT vs. Slovak-SNK), despite lacking access to overt morphological information. If case and word order traded off perfectly efficiently and case supplied all the relevant information, then we would have expected the case-marked models to perform nearly perfectly and the non-case-marked models to perform at chance.

Consistent with recent findings by Levshina (2021), these data therefore challenge the simple view that word order and morphology trade off since the benefits of the added grammatical complexity associated with any formal marking (word order / case marking / agreement rules) appear to be limited.

The presence of redundant marking can be explained by a number of factors. First, the simplest justification for redundancy in any communication code is the presence of noise in the transmission and receipt of signals (Shannon, 1948). Redundancy allows information to be recovered even in the presence of signal loss. Given that transmission in linguistic exchanges is often lossy, redundancy plausibly makes linguistic communication more robust to noise in terms of conveying its intended message (e.g., Aylett & Turk, 2004; Fenk-Oczlon & Fenk, 2008; Gibson, Bergen, et al., 2013; Jaeger, 2010; Levy, 2008; Wit & Gillette, 1999). And redundancy likely makes learning easier (Tal & Arnon, 2022).

Another possible justification for the existence of word order constraints has to do with increasing efficiency on the side of the language producer (e.g., see MacDonald, 2013). Language production is a complex cognitive feat, where a producer must select some words from among tens of thousands of words in their active vocabulary and combine them appropriately to convey some intended meaning. Producers are often faster when they are faced with fewer choices: objects for which multiple labels are possible (e.g., couch, futon, sofa) are slower to name than objects for which only one possible name exists (Lachman, 1973; Torrance et al., 2018). This phenomenon is an instance of a more general pattern where human choice behavior is slower when there are more options (Hick, 1952; Hyman, 1953). Rigid word order rules imply that the order of words in a sentence is fully determined by their grammatical roles, thus reducing the number of choices a speaker must make. This explanation is complicated, however, by findings indicating that language production is sometimes faster when there are more choices for the following syntactic construction (Ferreira, 1996; Ferreira & Dell, 2000; among others).

Finally, we consider the role of grammatical marking from a functional perspective. In language, being right *most of the time* may not be good enough, and the small number of sentences where, absent formal cues, the meaning is ambiguous are sufficient to give rise to regularized grammatical rules. Such cases include i) semantically reversible events where the two nominals both denote plausible subjects, typically clauses with two animate entities (e.g., Ray helped Lu /

Lu helped Ray), and ii) events that are unusual, i.e., violate the statistics of the world (e.g., the man bit the dog; cf. the more common event of the dog biting the man). Both instances occur often enough (sentences with animate subjects and animate objects occur ~10% of the time in our English sample; sentences with inanimate subjects and animate objects ~5% of the time) that there seems to be a functional benefit to being able to handle them in the grammar of a language. Moreover, the ability to grammatically identify the subject in sentences with animate subjects and objects may be a particularly important capacity since "humans like to talk about humans" (MacWhinney, 1977; Everett, 2009). And being able to say implausible things like *"man bit dog"* is a hallmark of language that allows for several of its most celebrated design features (Hockett, 1960), such as prevarication (lying) and displacement (talking about things that are not present or that do not even exist).

Although these cases are relatively rare, word order cues would only work if used *consistently*, even if they are *usually* redundant with word meanings. Otherwise, word order would not be reliable and thus not useful. For example, imagine a linguistic system in which the word order is SVO 70% of the time and OVS 30% of the time. A speaker wishing to convey an implausible sentence like *"man bit dog"* would be able to say either *"man bit dog"* or *"dog bit man."* A rational language producer, knowing that SVO is more common, might use the word order SVO in hopes that the comprehender would infer that *man* was the subject (since subjects usually precede objects in this hypothetical language). However, given that the prior probability of the utterance would be highly biased towards *"dog bit man,"* the comprehender would still be likely to infer that the intended meaning was *"man bit dog"*. On the other hand, if the language categorically used SVO order and categorically excluded OVS order, then *"man bit dog"* would be interpreted with 'man' as the subject and 'dog' the object, despite the implausibility of the resulting meaning.

The same logic does not apply to case or agreement marking because these cues, unlike word order, can be optional. For instance, one could imagine an efficient linguistic system in which case marking was not required to convey the plausible meaning *"dog bit man"*, but *was* required if one wanted to convey the implausible meaning *"man bit dog"*. In fact, differentially marking non-prototypical objects (e.g, human or animate objects) is a relatively common phenomenon across languages (e.g., in Spanish, specific human objects are marked by a preceding *a*, whereas most other objects are not), called differential object marking (Aissen, 2003). Therefore, for case or agreement marking to be a reliable cue, it does not need to always be present, unlike word order.

This account offers a possible explanation for why languages like English have relatively strict word orders even though, as our experiments show, most meanings can be inferred from word meanings alone. Were the word order not strict even in redundant instances, it would not be a sufficiently strong cue for overriding the plausibility of the meaning conveyed when needed.

That is, without strict word order, it would be impossible to say things like "the bone chewed the dog."[3]

This finding may make sense of the seeming ability of large language models in NLP to perform well in the absence of word order information. Given that in most cases word order information is redundant with word meanings, it is plausible for the overall performance of a model trained on scrambled input to be high. Notably, though, our account predicts that such models would suffer in cases where word order information *is* crucial. Consistent with this finding, Papadimitriou et al. (2022) show that the model BERT seems to rely on a different process for categorizing subjects and objects in sentences that convey prototypical semantic meanings ("The dog chewed the bone.") compared to those that do not ("The bone chewed the dog.").

Relatedly, in human language processing, scrambling word order does not reduce neural responses in the language-selective network unless local semantic composition is blocked (Mollica, Siegelman et al., 2020). This result, with our findings, may have implications for the use of semantically reversible sentences in cognitive neuroscience (e.g., Berndt et al., 1996; Caramazza & Zurif, 1976; Richardson et al., 2010; Thothathiri et al., 2012). Such sentences are commonly used in language research. The rationale for their use is that such materials allow researchers to isolate morpho-syntactic demands from those associated with the processing of word meanings and plausibility information. However, we would encourage the language research community to not simply ignore the fact that comprehenders can usually infer propositional meanings based on word meanings alone.

Of course, there is more to language than just transitive clauses and so the results here are specifically about subject vs. object selection, which—given well-known semantic differences that characterize subjects vs. objects (Dixon, 1994)—might be particularly amenable to being interpreted absent grammatical cues. Furthermore, the actual semantic roles of words in sentences are fine-grained and graded: not all subjects are agents, and not all agents are equally agentive, an observation that has spurred research into fine-grained taxonomies of semantic roles (Dowty, 1989; 1991; Kako, 2006; Reisinger, 2015; White, 2017). In future work, we hope to be able to expand this approach, integrating it into the broader literature on how sentence-level word order can (or cannot be) determined from its lexical items or sets of concepts alone (e.g., Chang et al. 2008; Chang, 2009; Malkin et al. 2021), how word order conveys information about not only argument structure but also information structure (Clark & Clark, 1978; Ferreira & Yoshida, 2003), and richer ideas about the different roles that words can play semantically.

---

[3] In fact, even *with* strict word order as in modern English, there is evidence from the literature on noisy-channel sentence processing (Gibson, Bergen, et al., 2013; Gibson et al., 2016, 2017; Ryskin et al., 2020) that meaning-based priors (e.g., that dogs chew bones and not *vice versa*) can sometimes cause human comprehenders to assume that an error had occurred somewhere in production or comprehension and to override grammatical cues in favor of the more plausible utterance (e.g., assuming that, even though they heard "the bone chewed the dog," the intended meaning was "the dog chewed the bone").

## 5. Conclusion

We propose that explaining the quantitative level of grammatical redundancy in natural language, which appears to be consistent across languages, should be a central goal in functional linguistics. From an information-theoretic perspective, the redundancy of natural language is one of its most distinctive features. Characterizing and explaining this redundancy has the potential to elucidate the relationship between form and function and to clarify the pressures that shape human language.

## Acknowledgments

KM was supported by NSF Award 2139005. EG was supported by NSF Award 2020840. EF was supported by NIH awards R01-DC016607, R01-DC016950, and U01-NS121471 and research funds from the McGovern Institute for Brain Research and the Simons Center for the Social Brain. We thank Zach Mineroff for help with setting up the English experiments, Yura Osadshii for help with recruiting Russian participants, Nafisa Syed for help with checking the English materials, and Inbal Arnon, Adele Goldberg, Ray Jackendoff, Isabel Papadimitriou, and members of Tedlab and Evlab for helpful discussions. We also thank our editor Franklin Chang and four anonymous reviewers.